\documentclass[10pt,twocolumn,letterpaper]{article}

\usepackage{cvpr}              % To produce the CAMERA-READY version
\usepackage{graphicx}
\usepackage{amsmath}
\usepackage{amssymb}
\usepackage{booktabs}
\usepackage{balance}
\usepackage[pagebackref,breaklinks,colorlinks]{hyperref}

\usepackage[capitalize, nameinlink]{cleveref}
\crefname{section}{Sec.}{Secs.}
\Crefname{section}{Section}{Sections}
\Crefname{table}{Table}{Tables}
\crefname{table}{Tab.}{Tabs.}

\def\cvprPaperID{20} % *** Enter the CVPR Paper ID here
\def\confName{CVPM}
\def\confYear{2022}

\def\slapnicar{Slapni\v{c}ar }

\begin{document}

%%%%%%%%% TITLE - PLEASE UPDATE
\title{Regression or Classification? Reflection on BP prediction from PPG data using Deep Neural Networks in the scope of practical applications}

\author{Fabian Schrumpf\\
Leipzig University of Applied Sciences\\
{\tt\small fabian.schrumpf@htwk-leipzig.de}
% For a paper whose authors are all at the same
% institution, omit the following lines up until the closing ``}''.
% Additional authors and addresses can be added with ``\and'', just like the
% second author.
% To save space, use either the email address or home page, not both
\and Paul Rudi Serdack\\
Leipzig University of Applied Sciences\\
{\tt\small paul\_rudi.serdack@stud.htwk-leipzig.de}
\and Mirco Fuchs\\
Leipzig University of Applied Sciences\\
{\tt\small mirco.fuchs@htwk-leipzig.de}
\thanks{This research was funded in part by the German Federal Ministry of Economics and Technology
(BMWi) (FKZ 49VF170043). The study at the Leipzig University Medical Center was
conducted according to the guidelines of the Declaration of Helsinki, and approved by the Ethics
Committee of the University of Leipzig (protocol code: 170/19-ek; date of approval: 6 July 2019).}
}

\maketitle

%%%%%%%%% ABSTRACT
\begin{abstract}
   Photoplethysmographic (PPG) signals offer diagnostic potential beyond heart rate analysis or blood oxygen level monitoring. In the recent past, research focused extensively on non-invasive PPG-based approaches to blood pressure (BP) estimation. These approaches can be subdivided into regression and classification methods. The latter assign PPG signals to predefined BP intervals that represent clinically relevant ranges. The former predict systolic (SBP) and diastolic (DBP) BP as continuous variables and are of particular interest to the research community. However, the reported accuracies of BP regression methods vary widely among publications with some authors even questioning the feasibility of PPG-based BP regression altogether. In our work, we compare BP regression and classification approaches. We argue that BP classification might provide diagnostic value that is equivalent to regression in many clinically relevant scenarios while being similar or even superior in terms of performance. We compare several established neural architectures using publicly available PPG data for SBP regression and classification with and without personalization using subject-specific data. We found that classification and regression models perform similar before personalization. However, after personalization, the accuracy of classification based methods outperformed regression approaches. We conclude that BP classification might be preferable over BP regression in certain scenarios where a coarser segmentation of the BP range is sufficient.
\end{abstract}

%%%%%%%%% BODY TEXT
\section{Introduction}
Predicting BP from single sensor signals such as PPG has gained a lot of attraction in recent years \cite{ding_continuous_2016, maqsood_survey_2022}. Using PPG is particularly interesting not only because sensors are cheap and easy to apply but also because the technique is related to remote PPG (rPPG; or imaging PPG / iPPG). RPPG generally refers to camera/video based derivation of PPG signals which allows to conduct remote measurements without any physical contact. If a reliable prediction of BP from PPG would be possible, there is reason to belief that concepts could be transferred and expanded to rPPG based BP prediction, which in fact has already been approached in several studies \cite{zou_non-contact_2021,souza_video-based_2021,takahashi_non-contact_2020}.

From a machine learning perspective, current approaches that use PPG for BP prediction can broadly be categorized into approaches based on extracting hand-crafted features \cite{tanveer_cuffless_2019, haddad_continuous_2020, el_hajj_cuffless_2020} and approaches that employ the entire signal and sometimes also its derivatives \cite{paviglianiti_neural_2020, Slapnicar2019}. The latter are usually based on a certain deep neural network (DNN) architecture. In DNN, these signals or their spectral representations are then usually either used directly in an end-to-end learning scheme to predict BP from the shape information \cite{paviglianiti_neural_2020} or are transformed into a spectrogram beforehand \cite{wu_improving_2021}. Even hybrid approaches have already been investigated \cite{Slapnicar2019, baek_end--end_2019}.

With respect to the target variable, approaches can be categorized into BP classification and BP regression. Classification is usually restricted to scenarios where authors are interested in predicting hypotension or hypertension versus normal BP \cite{cano_application_2021,sun_using_2021}. Regression, on the other hand, predicts BP as a continuous variable. With respect to the evaluation, usually only a mean error for the whole dataset along with a standard deviation or a some related metric is reported \cite{mahmud_shallow_2022,leitner_personalized_2021, jeong_combined_2021}. Some of these results imply suitability of the reported methods for medical applications since the errors reported are well below or at least close to the requirements of medical standards \cite{jeong_combined_2021, baek_end--end_2019}. In fact, a recent study has shown that much larger errors must be expected for a wide range of DNN architectures if the true underlying BP is not within the normotensive BP range \cite{schrumpf_assessment_2021}.

Assessing the reported results is often not straightforward. A reliable conclusion is usually difficult mainly due to issues with the underlying dataset. Main problems are (1) the
rather unbalanced nature of the underlying datasets which might lead to overfitting towards the mean of the distribution \cite{schrumpf_assessment_2021}, (2) issues with personalization within the datasets rendering the methods prone to overfitting to individual subjects \cite{kido_discussion_2021} and (3) the use of proprietary data collected in an experimental setting where the BP usually lacks the necessary variation within its range of interest. These varying boundary conditions render comparisons among different studies difficult especially if those studies try to solve different problems (i.e. BP classification and BP regression).

It stands to reason that well performing regression methods would be particularly interesting for medical applications. This would however require a robust performance within the full range of physiologically meaningful BP values. On the other hand, some applications certainly do not need highly accurately predicted BP values but rather a reliable mapping to a certain BP interval which we will refer to as BP bin for the rest of this paper (e.g. \cite{cano_application_2021,sun_using_2021}). The method of dividing the BP range into suitable intervals (i.e. number and widths of the intervals) to achieve adequate accuracy for clinical purposes is still an open question.

It is conceivable that certain scenarios only require a binary classification (high vs. low BP). From a medical perspective a meaningful segmentation would be the subdivision into intervals representing hypotension (hypoT), normotension (NT) and hypertension (HT) resulting in a multi class classification approach. On the other end, a dense binning could consist of very narrow BP intervals (e.g. bins of width 1 mmHg). It is obvious that the latter case is more suited for a regression-based approach since the underlying sorted nature of the BP values exhibits a larger importance with respect to the predicted target variable compared to cases with only a small number of classes.

Given these very different perspectives on the underlying problem, the following questions can be raised : (i) Is there reason to believe that classification and regression approaches for BP prediction should be preferred one over another in certain scenarios? (ii) Is there a trade-off between approaching the problem using a classification approach or a regression approach and the desired discretization of the BP range of interest into bins? (iii) And more generally, are classification approaches perhaps generally more suitable for BP prediction with respect to realistic scenarios since they might provide a better generalization to the available datasets which come with the above mentioned issues?

In this study, we compare regression and classification approaches based on several DNN architectures for BP prediction. From an evaluation perspective and for comparison reasons, we treat the whole problem as a classification task. While we directly derive a predicted class in a classification approach, we manually assign the target variable to a predicted class in the regression approaches based on its location in a certain BP interval. We applied this scheme to several segmentations of the BP range (i.e. subdivision with differently sized BP bins/intervals). Given each segmentation, we carefully prepared datasets based on the MIMIC-III database which allowed us to prevent overfitting to individual subjects as well as to any of the incorporated BP intervals. That means we ensured that our training sets are balanced with respect to the particular range segmentation into multiple intervals.

Other studies reported a positive effect of personalizing the network prior to prediction with subject-specific data \cite{Slapnicar2019, leitner_personalized_2021, hong_deep_2021}. This means that the pretrained network is partly or completely retrained using some portion of data from the subject used for prediction. Following this idea we also investigated in this study whether this personalization scheme is more or less effective with respect to the segmentation of the target variable (BP range) in the underlying BP prediction task. There is reason to believe that personalization is particularly effective for narrow bins, i.e. in problems very much related to regression but becomes less important if only a low number of BP classes with wide intervals are used.

The original contributions of this paper are as follows. First, we provide a systematic comparison of classification and regression based approaches for the task of BP classification using several BP range segmentations. Second, we evaluate the effect of personalization of DNNs using subject-specific data with respect to the granularity of the segmentation of the BP range for both the regression and classification scenario. Moreover, we show that the findings hold true for several DNN architectures.

The remainder of this paper is as follows: \Cref{sec:related_works} gives an overview over existing work in the field of BP prediction and classification. \Cref{sec:methods} outlines the methods used for creating the dataset, describes the relevant neural architectures and how they were trained in each scenario as well as the performance measures used for evaluating the results. \Cref{sec:results} presents the results and draws comparisons between regression-based and classification based approach with and without personalization. \Cref{sec:discussion} assesses the results presented in the light of our initial hypotheses. We also draw conclusions as to if and under what circumstances a decision for or against regression or classification might be justified.

%-------------------------------------- ---------------------------------
\section{Related Works}
\label{sec:related_works}
\subsection{BP regression}
Existing literature regarding PPG-based BP prediction can be divided into
approaches that try to predict SBP as well as DBP as continuous variables (i.e.
regression) and approaches that aim at classifying PPG signals into a set of
predefined classes (i.e. classification). The definition of these classes is
usually targeted towards the detection of certain adverse medical conditions
(e.g. hypotension (HypoT), normortension (NT), hypertension (HT)). Since
training of very deep neural architectures became feasible on consumer-grade
hardware, the attention of the research community has shifted mainly towards
regression-based approaches.

Regression-based approaches can be further subdivided into methods using
parameterized models or PPG features as well as end-to-end approaches.
Parameterized methods use the pulse transit time (PTT) or pulse arrival time
(PAT) to derive the pulse wave velocity (PWV). Several studies used linear
regression models to infer BP values from PWV \cite{Gesche2012, wippermann_evaluation_1995, zhang_pulse_2011}

 Feature-based approaches derive time and frequency domain features from PPG
 signals and use them as input to neural networks (NNs). Recently, Mahmud et al.
 \cite{mahmud_shallow_2022} designed a U-Net architecture that was tasked with deriving informative
 features from PPG and electrocardiography (ECG) signals. These features were then used to predict SBP
 and DBP using several clustering algorithms (e.g. kNN approaches and support
 vector machines). They achieved a MAE of 2.33 mmHg for SBP and 0.713 mmHg for
 DBP, respectively. Other authors designed recurrent NNs and derived
 time- and frequency-based features from PPG and ECG for predicting BP \cite{senturk_non-invasive_2020, jarm_novel_2021, harfiya_continuous_2021, el_hajj_cuffless_2020}.

 In contrast to feature-based methods, end-to-end approaches leverage the PPG waveforms themselves and implicitly derive informative features to predict BP. The advantage of this approach is that the selection and derivation of specific features is not necessary leaving the task of detecting patterns in the input data that are correlated to BP entirely to the neural architecture. Aguet et al. \cite{aguet_feature_2021} trained a siamese NN. They averaged consecutive PPG time windows to improve feature robustness to create the datasets for training and testing. Together with a callibration measurement the siamese network achieved an mean error of 0.31 mmHg for SBP and 0.4 mmHg for DBP, respectively. However, the standard deviation of these errors was high (10.27 mmHg for SBP and 5.62 mmHg for DBP). Leitner et al. \cite{leitner_personalized_2021} designed a hybrid neural architecture consisting of convolutional, recurrent and fully connected layers. After training with PPG signals, the NN was fine tuned using additional data from the test subjects. The authors achieved an MAE of 3.52 mmHg (SBP) and 2.2 mmHg (DBP). Jeong et al. \cite{jeong_combined_2021} used a convolutional neural network (CNN) in combination with a recurrent NN to predict BP. They processed PPG end ECG from the MIMIC-III database to create the dataset used for training and testing. Their architecture achieved a mean error of 0.02 mmHg (SBP) and 0.16 mmHg (DBP).

 \subsection{BP classification}
Many recent publications aiming at BP classification use already established neural architectures that are successfully applied to image classification tasks. Cano et al. \cite{cano_application_2021} modified pretrained GoogleNet and ResNet architectures and trained them using 50 subjects downloaded from the MIMIC-III database. The target variable was a classification into HT, pre-HT and NT. The highest F1-score on the test dataset was achieved with the ResNet18 network. Sun et al. \cite{sun_using_2021} used the Hilbert-Huang Transform on PPG signals and their first and second derivatives to fine tune a pretrained AlexNet architecture. The classification of the input signals into NT and HT achieved an accuracy score of 98.9 \%. Liang et al. \cite{liang_photoplethysmography_2018} computed the continuous wavelet transform (CWT) from the PPG signals of 121 records downloaded from the MIMIC-III database. A pretrained GoogLeNet was used to classify the scalograms into NT, pre-HT and HT. Multiple trainings were performed to invesitgate the accuracy when classifying the scalograms into pairwise combinations of the target classes. The highest F1-score of 92.55 \% was achieved when classifying NT and pre-HT.

Other authors created custom neural architectures for BP classification. Wu et al. \cite{wu_improving_2021} proposed a CNN designed for NT /HT classification based on the CWT of PPG signals. They achieved a validation accuracy of 90 \%. Mejía- Mejía et al. \cite{mejia-mejia_classification_2021} derived time and frequency domain features from the PPG-based pulse variablity signal. A subset among those features was selected based on a importance analysis. The authors trained various classification methods (e.g. k-NN, support vector machines and multilayer perceptrons) for HypoT/NT/HT classification. The highest accuracy on the test set was 70 \%.

%------------------------------------------------------------------------
\section{Methods}
\label{sec:methods}
\begin{figure}[t]
    \centering
    \includegraphics[width=0.8\columnwidth]{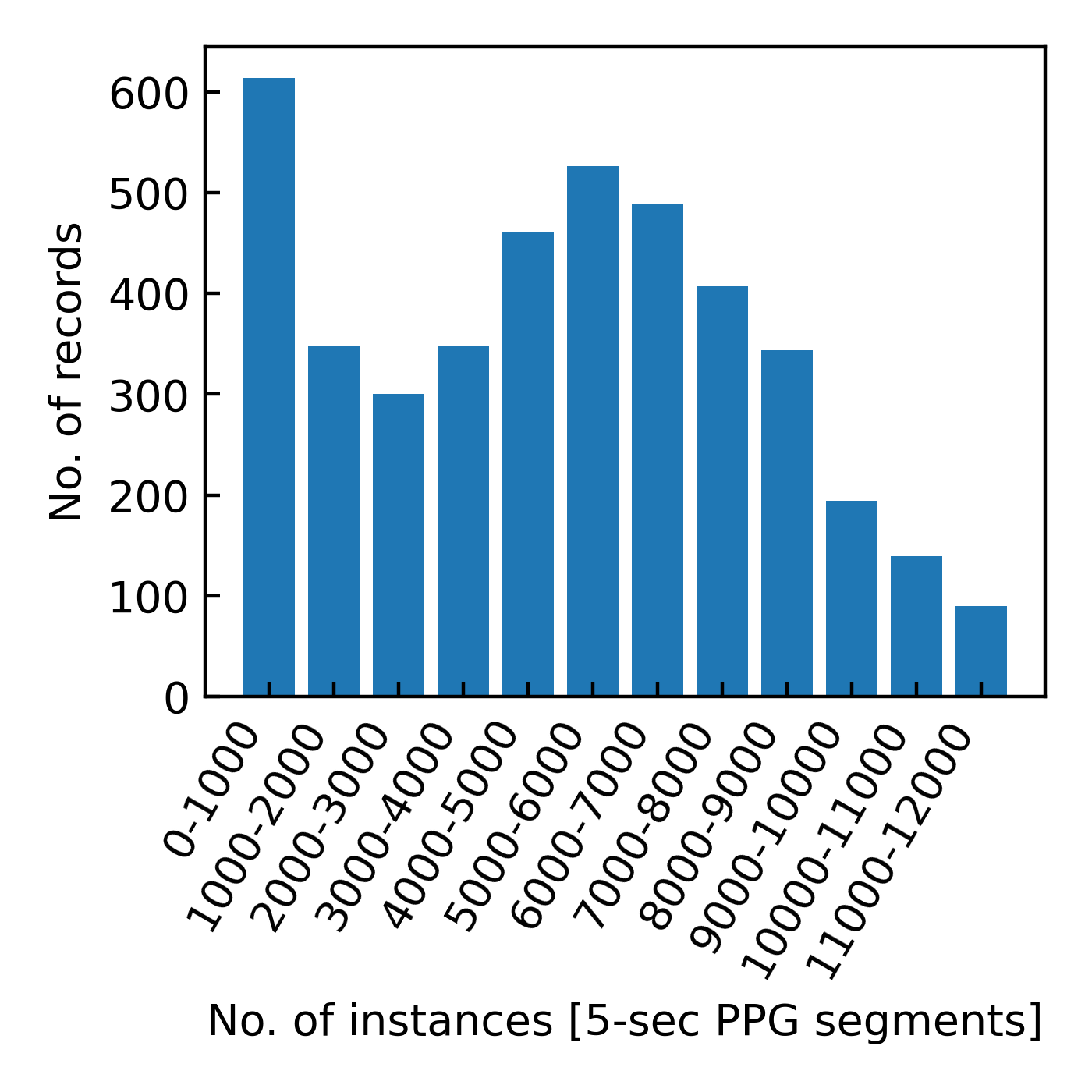}
    \caption{Histogram of the number of records downloaded from the MIMIC-III database containing a certain amount of PPG windows for training}
    \label{fig:HistSamplePerSubj}
\end{figure}
\subsection{Dataset}
The dataset used in this work is based on the MIMIC-III database \cite{johnson_mimic-iii_2016, Goldberger2000, johnson_alistair_mimic-iii_2015} and was created as described in \cite{schrumpf_assessment_2021}. Essential processing steps comprised downloading ABP and PPG signals from 4000 subjects off the MIMIC-III database using mining scripts provided by \slapnicar et al. \cite{Slapnicar2019}. The downloaded ABP and PPG data were divided into windows with a length of 5 s and an overlap of 2.5 s (50 \%) between consecutive windows. To create a balanced dataset, the samples had to be collected so that each subject contributed equally to the dataset. \autoref{fig:HistSamplePerSubj} shows a histogram of the number of records in the MIMIC-III database that contain a certain amount of samples. Since the majority of the records contribute at least 1000 samples to the dataset, only those records were selected for further processing. PPG signals were filtered using a 4th order Butterworth bandpass filter (\(f_{cut} = 0.5 Hz - 8Hz\)). Additionally, a quality check in terms of signal-to-noise ratio (SNR) was performed for every PPG window. Samples with an SNR below -7 dB were discarded. All PPG windows were normalized to zero mean and unit variance.

Ground truth SBP values were derived from the ABP signals. We selected the SBP as the sole target since it has proven to be a better indicator for cardiovascular risk than DBP \cite{strandberg2003most}. We employed a peak detection algorithm to detect the systolic peaks in each ABP window. The reference SBP was then derived by calculating the median of all SBP peaks in each signal window. To discard samples with a BP value outside the physiologically plausible range, we removed all samples with an SBP lower than 80 mmHg or higher than 180 mmHg. Signal windows with a median heart rate that exceeded the range of 50 to 140 bpm were also rejected.
\subsection{BP range segmentations}
\begin{figure}[t]
    \centering
    \includegraphics[width=0.95\columnwidth]{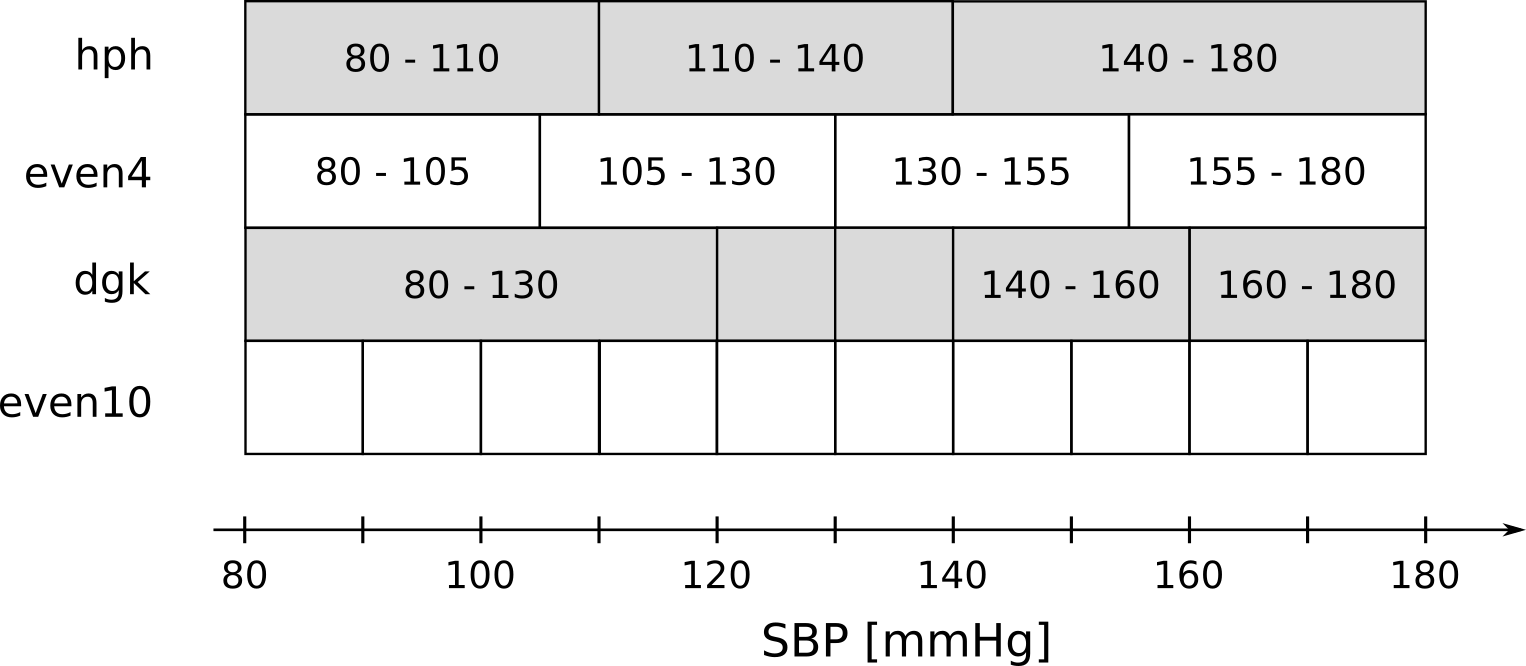}
    \caption{Overview over the BP range segmentation used for training the NNs. The SBP range was divided into variable numbers of bins.}
    \label{fig:binning}
\end{figure}
\begin{table}[t]
    \centering
    \begin{tabular}{ccc}
        \toprule
         BP range segmentation& No of subj, & No of samples\\
         \midrule
         hph & 1214& 3642000\\
         even4&475&1900000\\
         dgk&189&1134000\\
         even10&94&940000\\
        \bottomrule
    \end{tabular}
    \caption{Numbers of subjects and records included in the segmentations of the BP range used to train NNs. For details on the segmentations, see text. }
    \label{tab:SamplesPerSeg}
\end{table}
% \begin{table*}[t]
% \centering
% 	\begin{tabular}{lcccc}
% 		\toprule
% 		BP range classification&AlexNet&ResNet18&ResNet34&ResNet50\\
% 		\midrule
% 		\multicolumn{5}{l}{Classification}\\
% 		\midrule
% 		hph&0.45&0.44&0.45&0.44\\
% 		even4&0.36&0.36&0.37&0.36\\
% 		dkg&0.24&0.24&0.25&0.23\\
% 		even10&0.16&0.15&0.16&0.16\\
% 		\midrule
% 		\multicolumn{5}{l}{Regression}\\
% 		\midrule
% 		hph&0.42&0.46&0.45&0.45\\
% 		even4&0.36&0.36&0.37&0.38\\
% 		dkg&0.25&0.25&0.25&0.25\\
% 		even10&0.14&0.16&0.16&0.16\\
% 		\bottomrule
% 	\end{tabular}
% 	\caption{Test accuracy of the prediction performance of the neural networks for the different BP range segmentations under test. Results are reported separately for regression-based and classification-based approaches.}
% 	\label{tab:AccPretrain}
% \end{table*}
To investigate the influence of different segmentations of the BP range on the classification accuracy, we divided the SBP range into bins. Four different BP segmentation schemes shown in \autoref{fig:binning} were used. (1) \textit{hph}: 3 bins, their location and width reflects diagnostically meaningful BP ranges (i.e. hypoT, NT and HT) according to the German Cardiac Society (DGK) and the World Health Organization (WHO) \cite{williams_2018_2018, noauthor_guideline_2021}; (2) \textit{even4}: four equally sized bins covering the whole BP range; (3) \textit{dgk}: a division of the BP range into six physiologically meaningful intervals according to the DGK \cite{williams_2018_2018}; (4) \textit{even10}: 10 bins of width 10 mmHg, this approximates more regression-like approaches. For each BP segmentation we selected only subjects from the MIMIC-III database, whose BP value ranges spanned across all BP bins. In order to create balanced datasets, the contribution of each subject was limited to 1000 samples per bin. This led to datasets containing a variable number of subjects and samples depending on the number of bins in the dataset. An overview of the total number of subjects and samples for every BP segmentation can be found in \autoref{tab:SamplesPerSeg}.

\subsection{Neural network architectures}
Four different NN architectures were used for classification and regression. We used a modified version of the AlexNet architecture to classify BP into various bins \cite{krizhevsky_imagenet_2017}. Originally, AlexNet is a CNN that takes RGB images as input and classifies them into one of 1000 categories. We adopted the architecture for BP classification such that the first layer takes PPG time series data as input. Similarly, the number of output neurons of the final classification layer was adjusted according to the respective BP segmentation that was used for BP classification.

ResNets are very deep CNN consisting of blocks of convolutional layers with skip connections \cite{he_deep_2016}. These skip connections efficiently account for the vanishing gradient problem that occurs in very deep neural architectures \cite{srivastava_highway_2015}. We used three different versions of this architecture with varying depth. Specifically, we used ResNet18, ResNet34 and ResNet50.

The input dimensions of all networks were \(N_{samp} \times 1\) (1D PPG signal segment). Output dimensions of the final classification layers were \(N_{bin} \times 1\) for the classification problem, where \(N_{bin}\) is given by the number of bins in the particular BP range segmentation. In case of regression based-training the target variable corresponds to the SBP as a single value. Network weights of all models were initialized randomly using the Glorot method since it has proven to lead to a quicker convergence of the NN during training \cite{glorot_understanding_nodate}.

\subsection{NN training}
\subsubsection{Pretraining}
The data was split into chunks of 70 \%, 22.5 \% and 7.5 \% which were used for NN training, validation after each epoch and final testing. The split was achieved by assigning data from each subject to only one of these chunks. In comparison to a data split based on raw data samples this strategy prevents overfitting to subjects as data of particular subjects are not split across training, validation and test sets. Our approach to prevent an imbalance between the number of samples in the BP bins of the particular BP range segmentation is described in sec. 2.2.

Input and training pipelines as well as the neural architectures were implemented using Tensorflow 2.3 and Python 3.8. Training was performed using the Adam optimizer with an initial learning rate of 0.001 and a batch size of 128. The training was stopped if the validation loss stopped improving for 10 epochs and the best performing model was used for testing.

\subsubsection{Personalization}
Previous studies have shown that there are substantial differences in PPG morphology between different subjects that prevent a successful generalization of NNs across various patients \cite{Slapnicar2019}. Consequently, the prediction accuracy of NNs trained on data from an extensive subject population might be inadequate for clinical applications. Using subject-specific data to fine tune the pretrained NN has the potential to increase the prediction accuracy for individual subjects. To investigate the effect of personalization on the BP prediction accuracy for classification and regression based approaches we selected 10 patients from the MIMIC-III dataset that were previously used for testing the NNs. Each chosen subject's data was ordered by SBP value and every 10th sample was selected as data for fine tuning the NN. Since we ensured that the data of every subject spanned the whole SBP range during dataset creation, the data used for finetuning also fulfilled this criterion.

Our fine tuning strategy corresponds to a transfer learning approach \cite{leitner_personalized_2021} in which all weights of a pretrained network are allowed to be updated. The remaining data of the particular subject was split into equal parts and used for validating and testing the fine tuned NNs. Training was performed for a fixed number of 100 epochs. The model from the best epoch in terms of validation accuracy was used for testing.

\subsection{Evaluation metric}
We evaluated the performance of each NN in terms of accuracy with which the models predicted the correct BP bin. The accuracy metric is well suited since we ensured a balanced number of samples across classes. In the case of the BP regression, the predicted SBP value was assigned to its respective BP bin. This allowed a direct comparison between classification and regression evaluation results using classification-based metrics. Additionally, confusion matrices were calculated for a better grasp of the prediction characteristics.

%------------------------------------------------------------------------
\section{Results}
\label{sec:results}
\begin{figure*}[ht]
	\centering
	\includegraphics[width=0.9\textwidth]{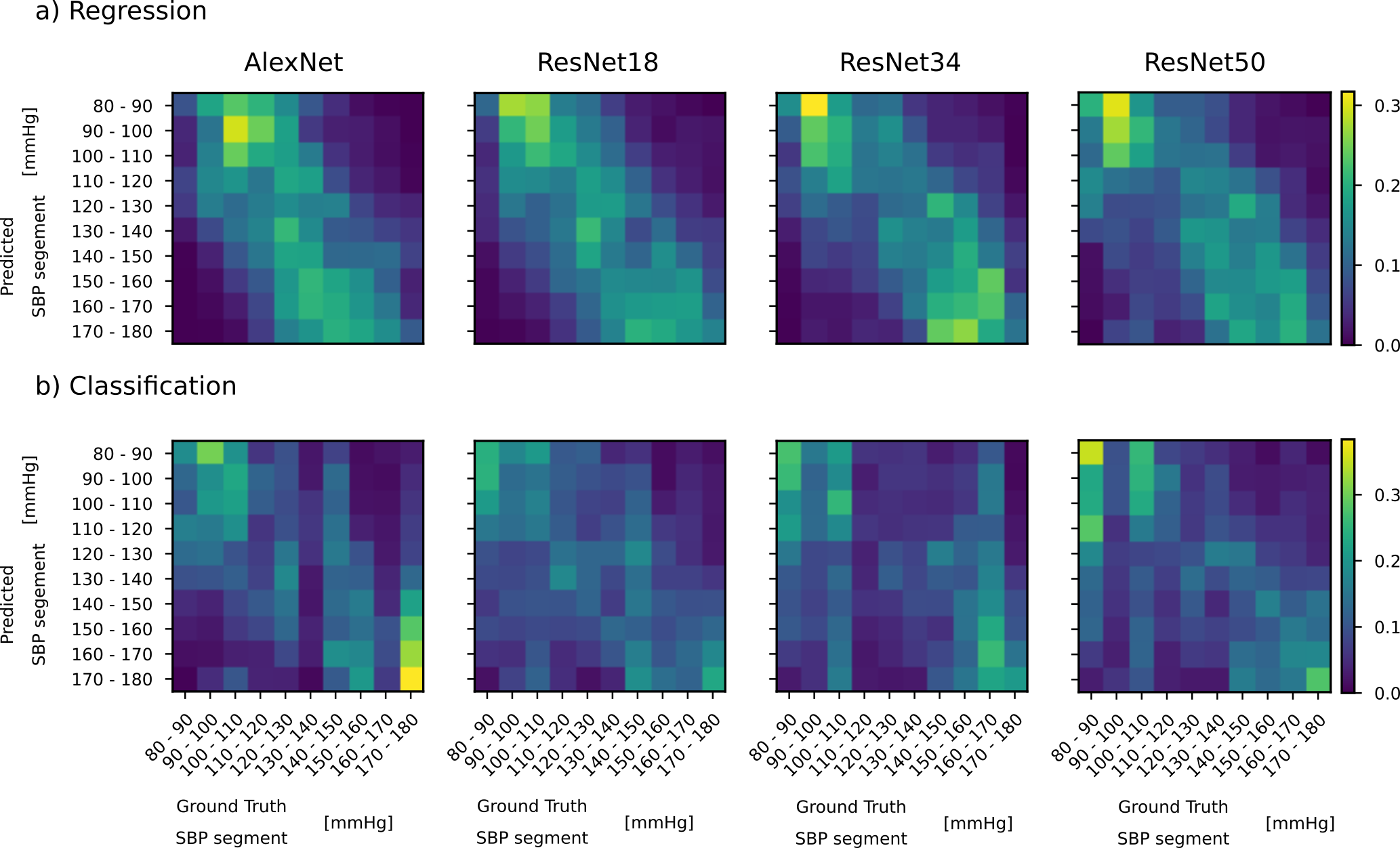}
	\caption{Confusion matrices for the pretraining of the neural architectures using the \textit{even10} segmentation. Ground truth BP classes and predicted BP classes are shown along the x- and y-axis respectively. Results are presented separately for regression-based (a) and classification-based (b) approaches. Number of samples in matrix element are normalized to the total number samples in the respective matrix row (i.e. number of samples in the respective class).}
	\label{fig:pretrain_confusion_CLA}
\end{figure*}
\begingroup
\setlength{\tabcolsep}{5pt} % Default value: 6pt
\renewcommand{\arraystretch}{0.9} % Default value: 1
\begin{table}[t]
\centering
	\begin{tabular}{lcccc}
		\toprule
		&AlexNet&ResNet18&ResNet34&ResNet50\\
		\midrule
		\multicolumn{5}{l}{Class.}\\
		\midrule
		hph&0.45&0.44&0.45&0.44\\
		even4&0.36&0.36&0.37&0.36\\
		dkg&0.24&0.24&0.25&0.23\\
		even10&0.16&0.15&0.16&0.16\\
		\midrule
		\multicolumn{5}{l}{Reg.}\\
		\midrule
		hph&0.42&0.46&0.45&0.45\\
		even4&0.36&0.36&0.37&0.38\\
		dkg&0.25&0.25&0.25&0.25\\
		even10&0.14&0.16&0.16&0.16\\
		\bottomrule
	\end{tabular}
	\caption{Test accuracy of the prediction performance of the NNs for the different BP range segmentations under test. Results are presented separately for regression-based and classification-based approaches. Predicted SBP values from the regression-based approach were assigned to their respective BP bin to allow for a direct comparison to the classification-based approach in terms of accuracy.}
	\label{tab:AccPretrain}
\end{table}
\endgroup
\autoref{tab:AccPretrain} shows the accuracy for every BP segmentation and every neural architecture on the test set after training the models from scratch. It can be seen that BP segmentations with just a small number of classes (\textit{hph}) achieve a higher accuracy than BP segmentations that divide the BP range into a bigger number of classes (\textit{dgk}). Comparing the different neural architectures, no significant difference in accuracy could be observed. Likewise, there was no difference in accuracy between the regression-based and the classification-based training.

\autoref{fig:pretrain_confusion_CLA} shows the confusion matrices for the regression-based and classification-based approaches using the \textit{even10} segmentation. The false detections indicated in the off-diagonal elements underpin the findings from the accuracy metrics in \autoref{tab:AccPretrain}. However, resulting matrices from the regression-based approaches show that most of the samples are grouped near the main diagonal suggesting that the prediction is in many cases only off by a few bins from the correct bin. This effect is most pronounced in the results from the \textit{even10} BP segmentation. The models seem to be more capable of distinguishing BP bins that are farther apart in the BP range than bins that are close to each other in comparison to classification models. This is not reflected in the accuracy measure which penalizes misclassification regardless of the proximity of the misclassified bin to the correct bin.

\subsection{Personalization}
% \begin{table*}[t]
% 	\centering
% 	\begin{tabular}{lcccccccc}
% 	\toprule
% 	&\multicolumn{2}{c}{AlexNet}&\multicolumn{2}{c}{ResNet18}&\multicolumn{2}{c}{ResNet34}&\multicolumn{2}{c}{ResNet50}\\
% 	\midrule
% 	&Class.&Reg.&Class.&Reg.&Class.&Reg.&Class.&Reg.\\
% 	\midrule
% 	\multicolumn{9}{l}{Before personalization}\\
% 	\midrule
% 	hph&0.3&0.4&0.3&0.4&0.4&0.4&0.3&0.3\\
% 	even4&0.3&0.3&0.3&0.3&0.3&0.3&0.3&0.3\\
% 	physio1&0.2&0.2&0.2&0.2&0.2&0.2&0.2&0.2\\
% 	even10&0.2&0.1&0.1&0.2&0.1&0.1&0.1&0.1\\
% 	\midrule
% 	After personalization\\
% 	\midrule
% 	hph&0.8&0.4&0.6&0.9&0.9&0.9&0.9&0.9\\
% 	even4&0.7&0.6&0.8&0.8&0.8&0.8&0.8&0.8\\
% 	physio1&0.7&0.5&0.8&0.7&0.8&0.7&0.7&0.7\\
% 	even10&0.6&0.4&0.6&0.6&0.7&0.6&0.6&0.6\\
% 	\bottomrule
% 	\end{tabular}
% 	\caption{Mean accuracy for all BP range segmentations and neural architectures befor and after personalization with subject-specific data}
% 	\label{tab:AccPersonalization}
% \end{table*}
\begin{figure*}
  \centering
  \includegraphics[width=0.85\textwidth]{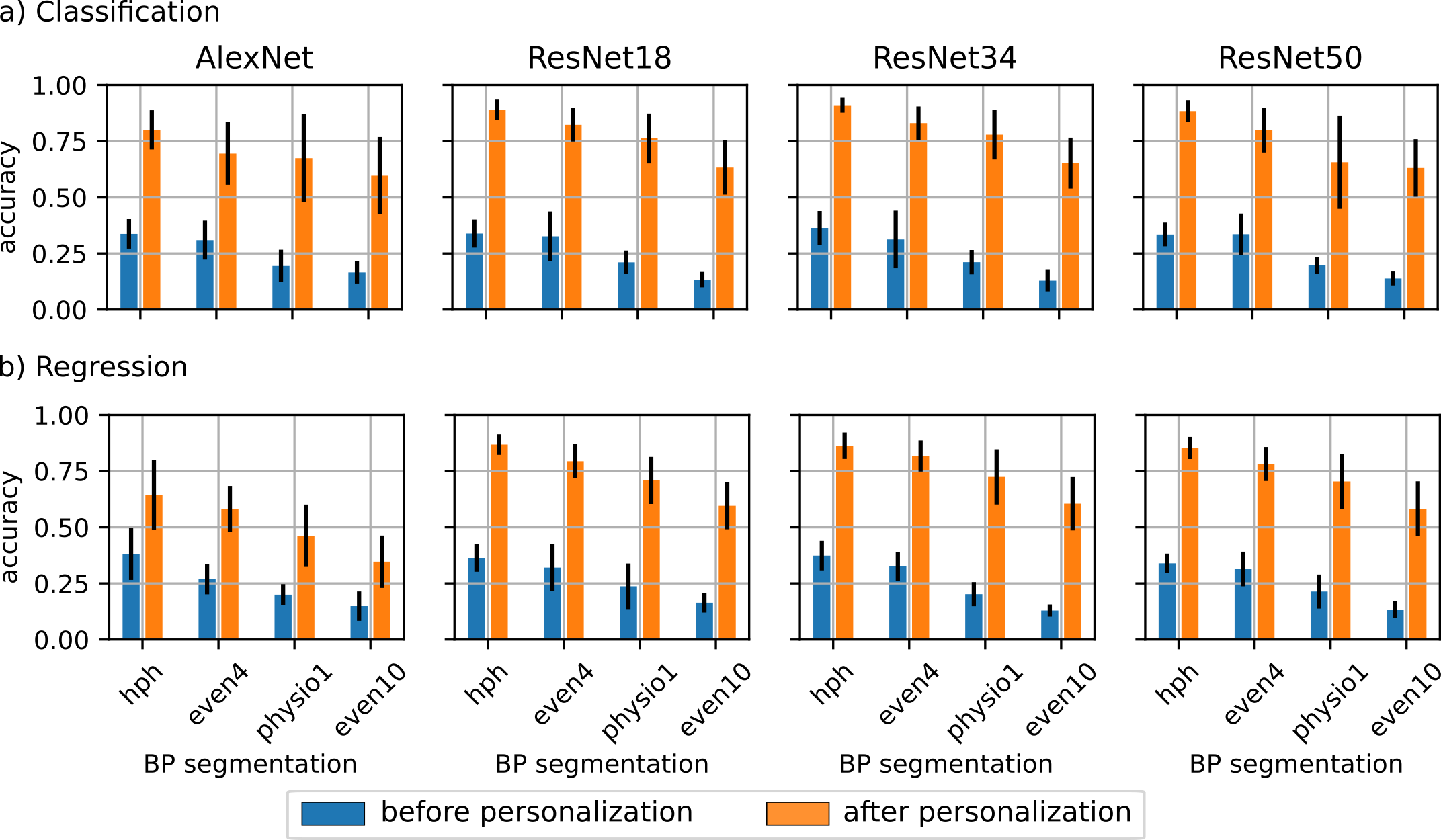}
  \caption{Mean and standard deviation of th test accuracy before (blue) and after (orange) personalization using subject-specific data of 10 test subjects for all considered BP segmentations and every neural architecture. Presented results are divided into results from the classification (a) and regression based (b) approaches.}
  \label{fig:acc_perso_bar}
\end{figure*}
\begin{figure}[]
  \centering
  \includegraphics[width=0.9\columnwidth]{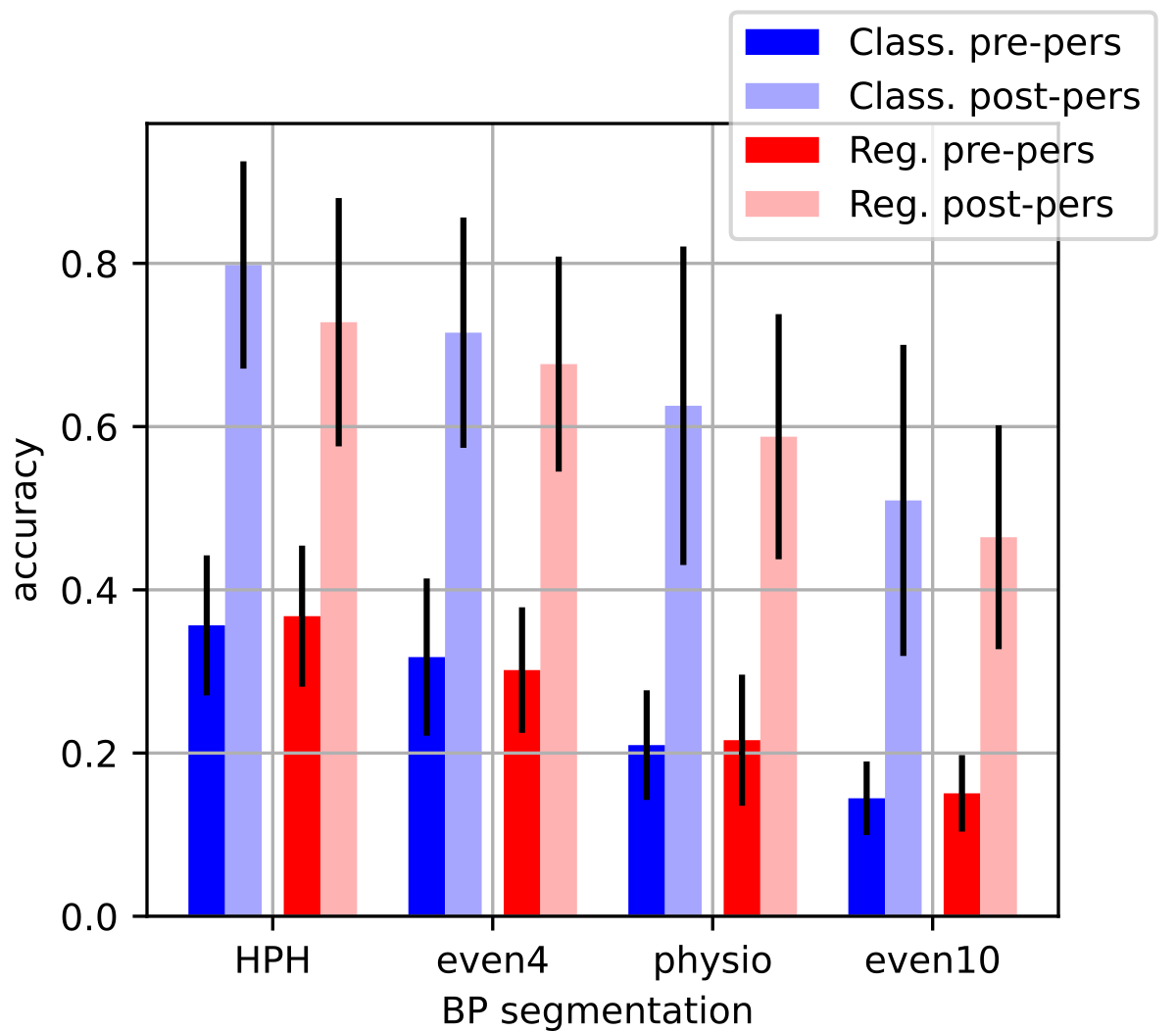}
  \caption{Mean and standard deviation of the test accuracy before and after personalization for all considered BP segmentations. Presented accuracies are averaged over all neural architectures.}
  \label{fig:acc_comp_binnings}
\end{figure}
Personalization was performed independently for each of the 10 selected subjects using 10 \% of the data. The remaining data was used in equal parts for validation and testing. We additionally verified the test performance on the original test set without the subject used for fine-tuning. This was done to ensure that the generalization capabilities of the network are maintained. Although the results are not shown here, we did not observe a significant drop of the performance after personalization. \autoref{fig:acc_perso_bar} shows the mean test accuracy for every training scenario. The models were evaluated before (blue bars) and after fine-tuning (orange bars). Test accuracy increased after fine tuning for every training scenario proving the efficacy of personalization for improving the performance on individual subjects. Similar to the results presented in \autoref{tab:AccPretrain} as well as \autoref{fig:pretrain_confusion_CLA} the test accuracy declined depending on the number of intervals used for the segmentation of the BP range. BP range segmentations with more bins resulted in a lower accuracy on the test set both pre- and post-personalization in comparison to segmentations with fewer bins. When comparing the various neural architectures, it can be seen that personalization had a greater effect on the ResNet variants as the increase in accuracy post-personalization seems to be higher in comparison to the AlexNet architecture.

Furthermore, we analyzed the differences in test accuracy between classification-based and regression based approaches for every BP range segmentation. Results are depicted in \autoref{fig:acc_comp_binnings}. We observed only small differences in pre-personalization accuracy when comparing classification- and regression-based accuracy. However, post-personalization accuracy was higher when predicting BP bins directly instead of performing a regression and assigning the appropriate BP bin to every predicted SBP-value. This effect seems to be mainly driven by the AlexNet results which show the most pronounced difference in post-personalization test accuracy between the regression and classification based scenario. However, a consistent minor effect can also be seen for all other network architectures. As stated before, test accuracy decreased with increasing numbers of BP bins.
%------------------------------------------------------------------------
\section{Discussion}
\label{sec:discussion}
This paper aimed at a comprehensive comparison of regression-based and classification-based non-invasive BP prediction using deep learning methods. We did not intend to derive a particularly accurate model to achieve state-of-the-art performance. Instead, we adopted a pragmatic approach to explore how the problem of PPG-based BP-prediction might be reformulated to answer relevant questions and be useful in a clinical setting. We derived an extensive dataset from the MIMIC-III database and trained well-established neural network architectures for both BP regression and classification. We divided the SBP into bins of varying number and width (BP range segmentations). The width and number of bins in these segmentations were designed to both cover physiologically relevant BP intervals (e.g. hypo-, normo- and hypertension) as well as to mimic a more regression-like approach (e.g. narrow bins of constant width). NN were trained using loss functions for regression and classification, while the evaluations were carried out with respect to classification scenarios.

Given this experimental setup, our objective was to answer several questions. We investigated whether classification or regression based approaches should be preferred one over another for BP class prediction in certain scenarios. More precisely, it might be plausible to prefer a classification loss during training for a coarse BP range segmentation and a regression loss in case of a denser segmentation. As our results indicate, this is not the case in general. It can be seen that the performance of both approaches constantly drops from coarser towards denser segmentations. This is obvious from the regression loss perspective since the predicted target variable (continuous BP values) is independent from the particular BP range segmentation. Therefore, the expected mean absolute error is the same for all segmentations which results in a higher chance of misclassifying the target towards denser segmentations, thus leading to a decrease of the performance.

Regarding the pre-personalization results there is no reason to prefer any of the two approaches for any of the given BP range segmentations, although regression methods seem to perform slightly better. This is different for post-personalization results where the classification based methods clearly outperform their regression counterparts. However, there is no indication that classifications based approaches might be less suited for denser BP segmentations compared to regression approaches. Given all of these results, there is also no obvious tradeoff between the coarseness of the BP range segmentation and the use of either regression or classification losses for training.

Our results also emphasize the importance of personalization, i.e. fine tuning the network weights with subject specific data. This personalization procedure, which has been proven effective in various studies before \cite{Slapnicar2019, leitner_personalized_2021, lee_beat--beat_2020}, leads to a strong performance increase for any network architectures and their training with both classification and regression losses compared to pre-personalization results. From the results in figure \ref{fig:acc_perso_bar} it can be seen that the ResNet architecture benefits the most from personalization. This may be due to the skip connections incorporated in the architecture that allow the network to converge to a better optimum compared to the AlexNet architecture. It seems to be even slightly more effective for classification based approaches since they outperform regression approaches after personalization which is not consistently the case before personalization. However, we could not find indications that the effectiveness of personalization depends on the particular task, i.e. the BP range segmentation.

We acknowledge that the subject population we used for personalization may not be sufficient to draw general conclusions. Additional selection criteria have to be employed to ensure that the subjects are representative for a population spanning a greater range of demographic and medical characteristics. Furthermore, it has to be investigated which properties and patterns enable the NN to improve its classification accuracy after training on subject-specific data. However, our results suggest that classification instead of regression has the potential to greatly improve the accuracy of non-invasive BP prediction.

Classification and regression of BP is subject to ongoing research and several relevant studies reported results that partly fulfilled the criteria of the British Hypertension Society and the Association of the Advancement of Medical Instrumentation. However, special attention has to be paid to the experimental setup and the design of the training pipeline in order to obtain unbiased results that allow for a realistic assessment of the method’s clinical applicability. In the light of these considerations, many authors question the practical feasibility of a truly continuous BP estimation \cite{schrumpf_assessment_2021}. One of the reasons may be external factors (e.g. age, chronical illnesses, medication and differences in measurement equipment) that introduce inter-subject variations into the PPG morphology which prevents the neural networks from good generalization. Given our experimental setup which ensured a balanced number of samples across BP bins in each segmentation, our post-personalization results indicate that it can be possible to achieve a BP classification performance of practical relevance solely based on PPG signals for some application scenarios.

Our results provide three insights: (i) training neural networks from scratch does not lead to an advantage in terms of test accuracy for either classification or regression approaches. Test accuracy drops as the number of BP segments gets larger. (ii) Personalization is immensely important for BP prediction as it enables machine learning methods to identify informative patterns in the subject-specific feature space through training on subject-specific data; (iii) classification-based approaches may be preferable over regression-based approaches when the correct association of BP to a small number of broad BP intervals is sufficient.

The findings in this work are of great importance when applying machine learning methods to camera-based PPG measurements. Such methods experience great interest from the scientific community and are investigated extensively \cite{zou_non-contact_2021,souza_video-based_2021,takahashi_non-contact_2020}. However, due to confounding factors like movement, reflections or changing lighting conditions that negatively impact the signal-to-noise ratio compared to their sensor based counterpart, it can be expected that the BP prediction error increases. Therefore, developing a method that satisfies all the relevant clinical criteria is still a work in progress. Simplifying the problem by classifying a limited number of BP ranges might be a way forward to arrive at a truly non-invasive BP prediction method that is applicable in a clinical setting.

%%%%%%%%% REFERENCES
{\small
\bibliographystyle{ieee_fullname}
\balance
\bibliography{CVPM_2022}
}

\end{document}